\documentclass[10pt,twocolumn,letterpaper]{article}

\usepackage[pagenumbers]{cvpr} 

\usepackage{graphicx}
\usepackage{amsmath}
\usepackage{amssymb}
\usepackage{booktabs}
\usepackage{xcolor}

\DeclareMathOperator*{\argminA}{arg\,min}
\usepackage{multirow}
\usepackage{makecell}
\usepackage[accsupp]{axessibility}

\newcommand{\quantize}{\mathbf{q}}
\newcommand{\Quantize}{\mathbf{q}}

\usepackage[pagebackref,breaklinks,colorlinks]{hyperref}

\usepackage[capitalize]{cleveref}
\crefname{section}{Sec.}{Secs.}
\Crefname{section}{Section}{Sections}
\Crefname{table}{Table}{Tables}
\crefname{table}{Tab.}{Tabs.}

\newcommand*{\affaddr}[1]{#1} 
\newcommand*{\affmark}[1][*]{\textsuperscript{#1}}
\newcommand*{\email}[1]{\texttt{#1}}
\usepackage[symbol]{footmisc}

\def\cvprPaperID{xxxx}
\def\confName{cvpr}
\def\confYear{2023}

\begin{document}

\title{CodeTalker: Speech-Driven 3D Facial Animation with Discrete Motion Prior}

\author{%
Jinbo Xing\affmark[1]~~ 
Menghan Xia\affmark[2,]$^*$~~ 
Yuechen Zhang\affmark[1]~~ 
Xiaodong Cun\affmark[2]~~ 
Jue Wang\affmark[2]~~ 
Tien-Tsin Wong\affmark[1]\\
\affaddr{\affmark[1]The Chinese University of Hong Kong~~~~~~~}
\affaddr{\affmark[2]Tencent AI Lab} \\
\small{\email{\{jbxing,yczhang21,ttwong\}@cse.cuhk.edu.hk ~~~~ \{menghanxyz,vinthony,arphid\}@gmail.com}}
}

\maketitle

\begin{abstract}
   Speech-driven 3D facial animation has been widely studied, yet there is still a gap to achieving realism and vividness due to the highly ill-posed nature and scarcity of audio-visual data.
Existing works typically formulate the cross-modal mapping into a regression task, which suffers from the regression-to-mean problem leading to over-smoothed facial motions.
In this paper, we propose to cast speech-driven facial animation as a code query task in a finite proxy space of the learned codebook, which effectively promotes the vividness of the generated motions by reducing the cross-modal mapping uncertainty.
The codebook is learned by self-reconstruction over real facial motions and thus embedded with realistic facial motion priors.
Over the discrete motion space, a temporal autoregressive model is employed to sequentially synthesize facial motions from the input speech signal, which guarantees lip-sync as well as plausible facial expressions. We demonstrate that our approach outperforms current state-of-the-art methods both qualitatively and quantitatively. Also, a user study further justifies our superiority in perceptual quality. Code and video demo are available at \url{https://doubiiu.github.io/projects/codetalker}.
\end{abstract}

\footnotetext[1]{~Corresponding Author.}

\section{Introduction}
\label{sec:intro}

3D facial animation has been an active research topic for decades, as attributed to its broad applications in virtual reality, film production, and games. The high correlation between speech and facial gestures (especially lip movements) makes it possible to drive the facial animation with a speech signal. Early attempts are mainly made to build the complex mapping rules between phonemes and their visual counterpart, which usually have limited performance~\cite{xu2013practical,taylor2012dynamic}. With the advances in deep learning,
recent speech-driven facial animation techniques push forward the state-of-the-art significantly. However, it still remains challenging to generate human-like motions.

As an ill-posed problem, speech-driven facial animation generally has multiple plausible outputs for every input. Such ambiguity tends to cause over-smoothed results.
Anyhow, person-specific approaches~\cite{karras2017audio,richard2021audio} can usually obtain decent facial motions because of the relatively consistent talking style, but have low scalability to general applications.
Recently, VOCA~\cite{cudeiro2019capture} extends these methods to generalize across different identities, however, they generally exhibit mild or static upper face expressions. This is because VOCA formulates the speech-to-motion mapping as a regression task, which encourages averaged motions, especially in the upper face that is only weakly or even uncorrelated to the speech signal.
To reduce the uncertainty, FaceFormer~\cite{fan2022faceformer} utilizes long-term audio context through a transformer-based model and synthesizes the sequential motions in an autoregressive manner. Although it gains important performance promotion, it still inherits the weakness of one-to-one mapping formulation and suffers from a lack of subtle high-frequency motions.
Differently, MeshTalk~\cite{richard2021meshtalk} models a categorical latent space for facial animation that disentangles audio-correlated and audio-uncorrelated information so that both aspects could be well handled.
Anyway, the employed quantization and categorical latent space representation are not well-suited for motion prior learning, rendering the training tricky and consequently hindering its performance.

We get inspiration from 3D Face Morphable Model (3DMM)~\cite{li2017learning}, where general facial expressions are represented in a low-dimensional space. Accordingly, we propose to formulate speech-driven facial animation as a code query task in a finite proxy space of the learned discrete codebook prior. The codebook is learned by self-reconstruction over real facial motions using a vector-quantized autoencoder (VQ-VAE)~\cite{van2017neural}, which along with the decoder stores the realistic facial motion priors. In contrast to the continuous linear space of 3DMM, combinations of codebook items form a discrete prior space with only finite cardinality. Still, in the context of the decoder, the code representation possesses high expressiveness.
Through mapping the speech to the finite proxy space, the uncertainty of the speech-to-motion mapping is significantly attenuated and hence promotes the quality of motion synthesis. Conceptually, the proxy space approximates the facial motion space, where the learned codebook items serve as discrete motion primitives.

Based on the learned discrete codebook, we propose a code-query-based temporal autoregressive model for speech-conditioned facial motion synthesis, called \textit{CodeTalker}. Specifically, taking a speech signal as input, our model predicts the motion feature tokens in a temporal recursive manner. Then, the feature tokens are used to query the code sequence in the discrete space, followed by facial motion reconstruction.
Thanks to the contextual modeling over history motions and cross-modal alignment, the proposed CodeTalker shows the advantages of achieving accurate lip motions and natural expressions.
Extensive experiments show that the proposed CodeTalker demonstrates superior performance on existing datasets. Systematic studies and experiments are conducted to demonstrate the merits of our method over previous works. 
The contributions of our work are as follows:
\begin{itemize}
    \item We model the facial motion space with discrete primitives in a novel way, which offers advantages to promote motion synthesis realism against cross-modal uncertainty.
    \item We propose a discrete motion prior based temporal autoregressive model for speech-driven facial animation, which outperforms existing state-of-the-art methods.
\end{itemize}



\section{Related Works}
\label{sec:related_works}

\subsection{Speech-driven 3D Facial Animation}

Computer facial animation is a long-standing task~\cite{parke2008computer} and has attracted rapidly increased interest over the past decades~\cite{weise2011realtime,li2013realtime,cao2016real,kim2018deep,zollhofer2018state,fried2019text,thies2020neural,lahiri2021lipsync3d}. 
As a branch, speech-driven facial animation is to reenact a person in sync with input speech sequences.
While extensive literature in this field works on 2D talking heads~\cite{chen2020talking,chen2018lip,chung2016out,das2020speech,ji2021audio,prajwal2020lip,vougioukas2020realistic,wang2022one,yi2020audio,zhou2021pose,liang2022expressive,ji2022eamm,alghamdi2022talking,liu2022semantic,shen2022learning,guo2021ad,pang2023dpe,zhang2022metaportrait}, we focus on facial animation on 3D models in this work, which can be roughly categorized into linguistics-based and learning-based methods.

\textbf{Linguistics-based methods.}
Typically, linguistics-based methods~\cite{massaro2012animated,taylor2012dynamic,xu2013practical,edwards2016jali} establish a set of complex mapping rules between phonemes and their visual counterparts, \ie, visemes~\cite{fisher1968confusions,lewis1991automated,mattheyses2015audiovisual}. For example, the dominance function~\cite{massaro2012animated} is to determine the influence of phonemes on the respective facial animation control parameters. Xu \etal~\cite{xu2013practical} defines animation curves for a constructed canonical set of visemes to generate synchronized mouth movements. There are also some methods considering the many-to-many mapping between phonemes and visemes, as demonstrated in the dynamic visemes model~\cite{taylor2012dynamic} and, more recently, the JALI~\cite{edwards2016jali}. Based on psycholinguistic considerations and built upon the Facial Action Coding System (FACS)~\cite{ekman1978facial}, JALI factors mouth movements into lip and jaw rig animation and generate compelling co-articulation results. Although these methods have explicit control over the animation, they have complex procedures and lack a principled way to animate the entire face.

\textbf{Learning-based methods.}
Learning-based methods~\cite{cao2005expressive,liu2015video,taylor2017deep,pham2018end,karras2017audio,cudeiro2019capture,wang20213d,habibie2021learning,fan2022faceformer,fan2022joint} resort to a data-driven framework. Cao \etal~\cite{cao2005expressive} achieve emotional lip sync by the proposed constrained search and Anime Graph structure. Recently, Taylor \etal~\cite{taylor2017deep} propose a deep-learning-based model utilizing a sliding window approach on the transcribed phoneme sequences input. Karras \etal~\cite{karras2017audio} propose a convolution-based network with a learnable emotion database to animate a speech-driven 3D mesh. More recently, VisemeNet~\cite{zhou2018visemenet} employ a three-stage Long Short-Term
Memory (LSTM) network to predict the animation curve for a lower face lip model. 

We review the most related works more concretely here as they have the same setting as this work, \ie, training on high-resolution paired audio-mesh data and speaker-independently animating entire face meshes in vertex space. MeshTalk~\cite{richard2021meshtalk} successfully disentangles audio-correlated and uncorrelated facial information with a categorical latent space. However, the latent space adopted is not optimal with limited expressiveness, thus the animation quality is not stable when applied in a data-scarcity setting. VOCA~\cite{cudeiro2019capture} employs powerful audio feature extraction models and can generate facial animation with different speaking styles. Furthermore, FaceFormer~\cite{fan2022faceformer} considers long-term audio context with transformer~\cite{vaswani2017attention} rendering temporally stable animations. Despite the appealing animations, both suffer from the over-smoothing problem, as they directly regress the facial motion in the highly ill-posed audio-visual mapping with large uncertainty and ambiguity.

\subsection{Discrete Prior Learning}
\label{subsec:discrete_prior_learning}

In the last decades, discrete prior representation with learned dictionaries has demonstrated its superiority in image restoration tasks~\cite{elad2006image,gu2015convolutional,timofte2013anchored,timofte2014a,jo2021practical}, since clear image details are well-preserved in the dictionaries. This line of techniques further inspires the high-capacity and high-compressed discrete prior learning. VQ-VAE~\cite{van2017neural} first presents to learn discrete representations (codebook) of images and autoregressively model their distribution for image synthesis. The follow-up works, VQ-VAE2~\cite{razavi2019generating} and VQ-GAN~\cite{esser2021taming} further improve the quality of high-resolution image synthesis. Recently, discrete prior learning has been exploited for image colorization~\cite{huang2022unicolor}, inpainting\cite{peng2021generating}, blind face restoration~\cite{zhou2022towards}, text-to-image synthesis~\cite{gu2022vector}, \etc.

In addition to image modality, most recent works also explore the power of discrete prior learning in tasks with other modalities, such as dyadic face motion generation~\cite{ng2022learning}, co-speech gesture synthesis~\cite{ao2022rhythmic}, speech enhancement~\cite{yang2022audio}. Inspired by codebook
learning, this work investigates to learn discrete motion prior for speech-driven 3D facial animation. Different from
~\cite{ng2022learning}, we exploit the discrete motion primitives for facial motion representation in a context-rich manner, which is more effective to learn general priors.

\vspace{-0.5mm}
\section{Method}
\label{sec:method}
\vspace{-0.5mm}



We aim to synthesize sequential 3D facial motions from a speech signal, so that any neutral face mesh could be animated as a lip-synchronized talking face. However, this is an ill-posed problem since one speech could be matched by multiple potential facial animations.
Such ambiguity tends to make cross-modal learning suffer from averaged motions and lack of subtle variations. To bypass this barrier, we propose to first model the facial motion space with the learned discrete motion prior, and then learn a speech-conditioned temporal autoregressive model over this space, which promotes robustness against the cross-modal uncertainty.
\vspace{-4mm}
\paragraph{Formulation.}
Let $\mathbf{M}_{1:T}=(\mathbf{m}_1,...,\mathbf{m}_T)$ be a sequence of facial motions, where each frame $\mathbf{m}_t \in \mathbb{R}^{V\times3}$ denotes the 3D movement of $V$ vertices over a neutral-face mesh template $\mathbf{h}\in \mathbb{R}^{V\times3}$. Let further $\mathbf{A}_{1:T}=(\mathbf{a}_1,...,\mathbf{a}_T)$ be a sequence of speech snippets, each of which $\mathbf{a}_t \in \mathbb{R}^{d}$ has $d$ samples to align with the corresponding (visual) frame $\mathbf{m}_t$. Then, our goal is to sequentially synthesize $\mathbf{M}_{1:T}$ from $\mathbf{A}_{1:T}$ so that an arbitrary neutral facial template $\mathbf{f}$ could be animated as $\mathbf{H}_{1:T}=\{\mathbf{m}_1+\mathbf{h},...,\mathbf{m}_T+\mathbf{h}\}$.




\subsection{Discrete Facial Motion Space}
\label{subsec:motion_space}
Visual realistic facial animations should present accurate lip motions and natural expressions. To achieve this from speech signals, extra motion priors are required to reduce the uncertainty and complement realistic motion components. As witnessed by the recent image restoration task~\cite{zhou2022towards}, discrete codebook prior~\cite{van2017neural} demonstrates advantages in guaranteeing high-fidelity results even from a severely degraded input.
Inspired by this, we propose to model the facial motion space as a discrete codebook by learning from tracked real-world facial motions.
\vspace{-4mm}
\paragraph{Codebook of motion primitives.}
We manage to learn a codebook $\mathcal{Z}=\{\mathbf{z}_k \in \mathbb{R}^{C}\}_{k=1}^{N}$ that allows any facial motion $\mathbf{m}_t$ to be represented by a group of allocated items $\{\mathbf{z}_k\}_{k \in \mathcal{S}}$,
where $\mathcal{S}$ denotes the selected index set through Eq.~\ref{eqn:quantization}.
Conceptually, the codebook items serve as the motion primitives of a facial motion space.
To this end, we pre-train a transformer-based VQ-VAE that consists of an encoder $E$, a decoder $D$, and a context-rich codebook $\mathcal{Z}$, under the self-reconstruction of realistic facial motions.
As shown in Figure~\ref{fig:motion_prior}, the facial motions $\mathbf{M}_{1:T}$ is first embedded as a temporal feature $\hat{\mathbf{Z}} = E(\mathbf{M}_{1:T})\in \mathbb{R}^{T' \times H \times C}$, where $H$ is the number of face components and $T'$ denotes the number of encoded temporal units ($P=\frac{T}{T'}$ frames). Then, we obtain the quantized motion sequence $\mathbf{Z}_\quantize \in \mathbb{R}^{T^\prime \times H \times C}$ via an element-wise quantization function $Q(\cdot)$ that maps each item in  $\hat{\mathbf{Z}}$ to its nearest entry in codebook $\mathcal{Z}$:

\begin{figure}[!t]
    \centering
    \includegraphics[width=1\linewidth]{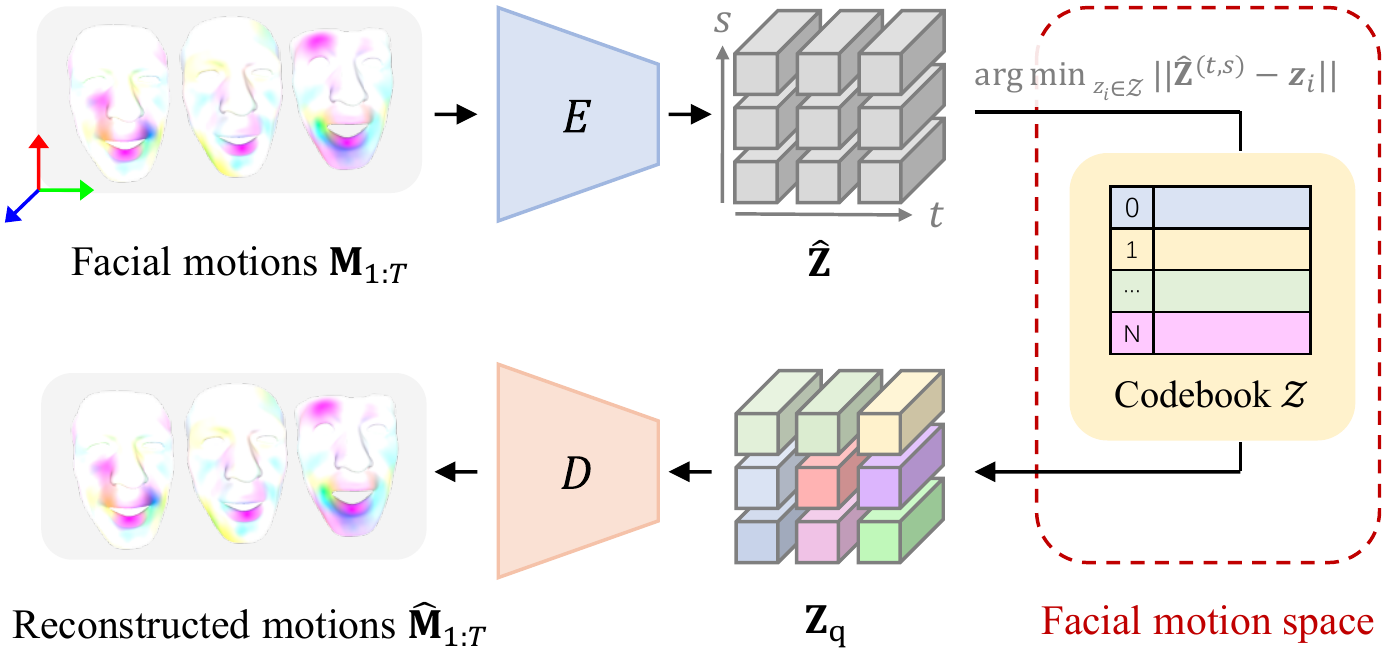}
    \caption{Learning framework of facial motion space. The learned motion primitives, as embedded in the codebook, serve to represent the facial motions in a spatial and temporal manner.}
    \label{fig:motion_prior}
    \vspace{-5mm}
\end{figure}

\begin{equation}
\mathbf{Z}_\quantize = Q(\hat{\mathbf{Z}}) := \argminA_{\mathbf{z}_k\in \mathcal{Z}} \Vert\hat{\mathbf{z}}_t-\mathbf{z}_k\Vert_2.
\label{eqn:quantization}
\end{equation}
Then, the self-reconstruction is given by:
\begin{equation}
\hat{\mathbf{M}}_{1:T} = D(\mathbf{Z}_\quantize) = D(Q(E(\mathbf{M}_{1:T}))).
\label{eqn:reconstruct}
\end{equation}
Note that, the discrete facial motion space reduces the mapping ambiguity with the finite cardinality, but never sacrifices its expressiveness thanks to its context-rich representation as a latent space.

\begin{figure*}[!t]
    \centering
    \includegraphics[width=1\linewidth]{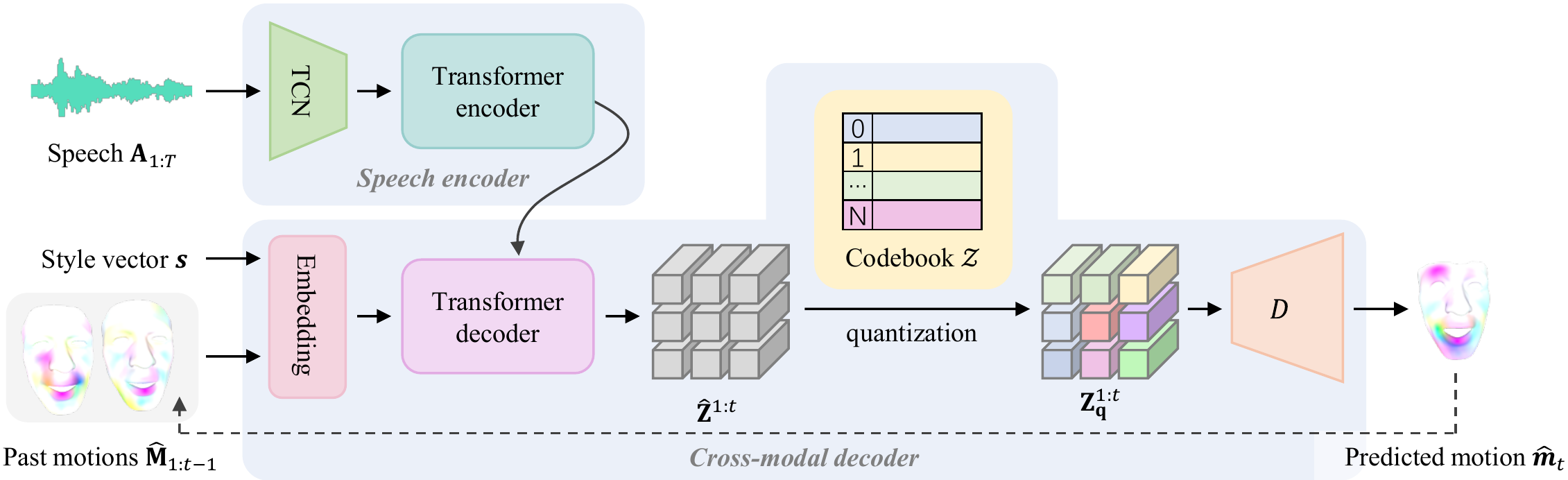}\vspace{-0.5em}
    \caption{Diagram of our speech-driven motion synthesis model. Given the speech $\mathbf{A}_{1:T}$ and style vector $\mathbf{s}$ as input, the model learn to recursively generate a sequence of facial motions by predicting the motion codes. As embedded with well-learned motion priors, the pre-trained codebook and decoder are frozen during training.}
    \label{fig:overview}\vspace{-1.5em}
\end{figure*}

\begin{figure}[!t]
    \centering
    \includegraphics[width=1\linewidth]{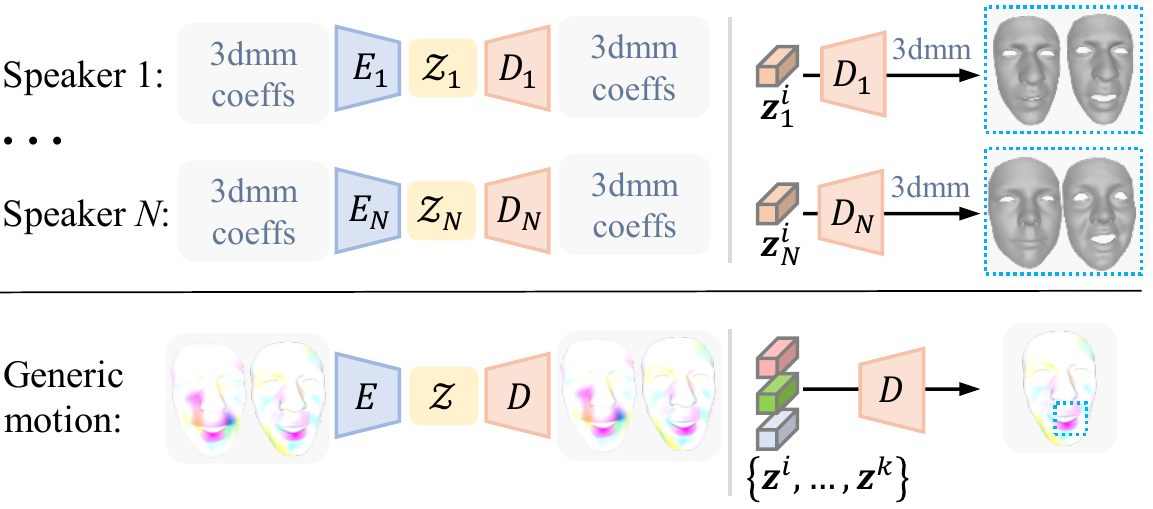}
    \caption{Concept comparison with Learn2Listen~\cite{ng2022learning}. (Top) The speaker-specific facial expression coefficient prior~\cite{ng2022learning}, in which each code represents a sequence of facial expression coefficients. (Bottom) Our speaker-agnostic generic motion prior, in which each code represents the motion primitive of face components. The blue dotted boxes indicate what information each code may represent conceptually.}
    \label{fig:motion_prior_discussion}
    \vspace{-6mm}
\end{figure}
\vspace{-4mm}
\paragraph{Training objectives.} Similar to \cite{van2017neural}, to supervise the quantized autoencoder training, we adopt a motion-level loss and two intermediate code-level losses:
\begin{equation}
\begin{aligned}
    \mathcal{L}_\text{VQ} = &\Vert \mathbf{M}_{1:T} -\hat{\mathbf{M}}_{1:T}\Vert_1 \\
    &+\Vert \text{sg}(\hat{\mathbf{Z}})-\mathbf{Z}_\quantize\Vert_2^2+\beta\Vert \hat{\mathbf{Z}}-\text{sg}(\mathbf{Z}_\quantize)\Vert_2^2,
\end{aligned}
\end{equation}
where the first term is a reconstruction loss, the latter two are adopted to update the codebook items by reducing the distance between the codebook $\mathcal{Z}$ and embedded features $\hat{\mathbf{Z}}$. sg($\cdot$) stands for a stop-gradient operation and $\beta$ is a weighting factor controlling the update rate of the codebook and encoder. Since the quantization function (Eq.~\ref{eqn:quantization}) is not differentiable, the straight-through gradient estimator~\cite{bengio2013estimating,van2017neural} is employed to copy the gradients from the decoder input to the encoder output.
\vspace{-4mm}
\paragraph{Discussion.}
Recently, Learn2Listen~\cite{ng2022learning} has applied VQ-VAE for facial expression synthesis in response to a given talking head harnessing 2D monocular videos to obtain 3DMM coefficients. In addition to distinct applications, here we would like to emphasize our major differences.
First, Learn2Listen constructs speaker-specific codebooks while ours uses a generic codebook that is feasible to represent arbitrary facial motions. Since cross-character motions are absorbed, our codebook is naturally embedded with more plentiful priors.
Second, Learn2Listen utilizes the codebook to represent common sequences of facial expressions by the way of 3DMM coefficients, \ie, each code represents a sequence (8 frames) of facial expressions. Differently, our codebook is formulated to represent the vertex-based facial motion space, where the codes are embedded with per-vertex motions of facial components and represent the facial motion~(within a temporal unit) in a context-rich manner.
As compared in Figure~\ref{fig:motion_prior_discussion}, the codebook of Learn2Listen is learned to memorize typical sequential facial expressions of a specific speaker within 3DMM space, which cannot synthesize realistic facial motions with subtle details due to the limited expressiveness of 3DMM and is bounded by the accuracy of 3D reconstruction techniques~\cite{ng2022learning}. As the first attempt, our codebook is learned to represent the generic facial motion space with motion primitives for captured facial mesh data, which is more effective to embed general priors preserving vivid facial details.

%

We further discuss the hyper-parameters of the codebook. First, the length of the temporal unit $P$ and the number of face components $H$ determine the complexity of the motion primitives in temporal and spatial aspects respectively. Generally, complex motion primitives cause low flexibility and reusability and thus hinder representation effectiveness. On the opposite, overly simple motion primitives challenge motion prediction due to the lack of semantics. Besides, the codebook size $N$ and the feature dimension $C$ determine the representation capability, which should be defined according to the complexity of the dataset and in cooperation with $P$ and $H$.
In our experiment, we set $N=256$, $P=1$, $H=8$ or $H=16$, and $C=64$ or $C=128$ depending on the dataset, which lead to high-quality results as justified by the ablation studies in Section~\ref{subsec:ablation}. More details can be found in the \emph{Supplement}.

\subsection{Speech-Driven Motion Synthesis}
\label{subsec:motion_prediction}

With the learned discrete motion prior, we can build a cross-modal mapping from the input speech to the target motion codes that could be further decoded into realistic facial motions. Along with the speech, we further adopt a control on the talking styles as input, \ie, a style vector $\mathbf{s} \in \mathbb{R}^M_+ \cup \{0\}$, where $M$ is the dimension of the learned style space (see Eq.~\ref{eqn:embedding}). Conditioning on the speech $\mathbf{A}_{1:T}$ and the style vector $\mathbf{s}$, a temporal autoregressive model, composed of a speech encoder $E_{\text{speech}}$ and a cross-modal decoder $D_{\text{cross-modal}}$, is employed to learn over the facial motion space, as depicted in Figure~\ref{fig:overview}.

Following FaceFormer~\cite{fan2022faceformer}, our speech encoder adopts the architecture of the state-of-the-art self-supervised pre-trained speech model, wav2vec 2.0~\cite{baevski2020wav2vec}, which consists of an audio feature extractor and a multi-layer transformer encoder. The audio feature extractor converts the speech of raw waveform into feature vectors through a temporal convolutions network (TCN). Benefiting from the effective attention scheme, the transformer encoder converts the audio features into contextualized speech representations.
Apart from the pre-trained codebook and VQ-VAE decoder, our cross-modal decoder contains an embedding block and a multi-layer transformer decoder with causal self-attention.
The embedding block combines the past facial motions and the style embedding via:
\begin{equation}
\mathbf{F}_\text{emb}^{1:t-1} = \mathcal{P}_\theta(\hat{\mathbf{M}}_{1:t-1}) + \mathbf{B} \cdot \frac{\mathbf{s}}{\Vert\mathbf{s}\Vert_1},
\label{eqn:embedding}
\end{equation}
where $\mathcal{P}_\theta$ is a linear projection layer, and $\mathbf{B} = [\mathbf{b}_1,...,\mathbf{b}_M]\in \mathbb{R}^{C \times M}$ denotes the $M$ learnable basis vectors that span the style space linearly. Alike to FaceFormer~\cite{fan2022faceformer}, we equip the transformer decoder with causal self-attention to learn the dependencies between each frame in the context of the past facial motion sequence, and with cross-modal attention to align the audio and motion modalities. The output features $\hat{\mathbf{Z}}^{1:t}$ is further quantized into $\mathbf{Z}^{1:t}_\quantize$ via Eq.~\ref{eqn:quantization} and decoded by the pre-trained VQ-VAE decoder. The newly predicted motion $\hat{\mathbf{m}}_t$ is used to update the past motions as $\hat{\mathbf{M}}_{1:t}$, in preparation for the next prediction. Formally, this recursive process can be written as:
\begin{equation}
    \hat{\mathbf{m}}_t = D_{\text{cross-modal}}(E_{\text{speech}}(\mathbf{A}_{1:T}), \mathbf{s}, \hat{\mathbf{M}}_{1:t-1}).
\label{eqn:prediction}
\end{equation}
\vspace{-9mm}
\paragraph{Training objectives.}
We train the transformer encoder, decoder and the embedding block for cross-modality mapping, while keeping the codebook $\mathcal{Z}$ and motion decoder $D$ frozen. To benefit from the speech representation learning from large-scale corpora, we initialize the TCN and transformer encoder with the pre-trained wav2vec 2.0 weights. Overall, the autoregressive model is trained in a teaching-forcing scheme, under the constraint of two loss terms:
(i) feature regularity loss $\mathcal{L}_\text{reg}$ measuring the deviation between the predicted motion feature $\hat{\mathbf{Z}}^{1:T}$ and the quantized feature $\mathbf{Z}^{1:T}_\quantize$ from codebook, and
(ii) motion loss $\mathcal{L}_\text{motion}$ measuring the difference between the predicted motions $\hat{\mathbf{M}}_{1:T}$ and the ground-truth motions $\mathbf{M}_{1:T}$, which plays an important role to stabilize the training process.
The final loss function is:
\begin{equation}
\begin{aligned}
\mathcal{L}_\text{syn} &= \mathcal{L}_\text{reg}  + \mathcal{L}_\text{motion} \\
        &= \Vert \hat{\mathbf{Z}}^{1:T}-\text{sg}(\mathbf{Z}^{1:T}_\quantize) \Vert_2^2 + \Vert \hat{\mathbf{M}}_{1:T} - \mathbf{M}_{1:T} \Vert_2^2.
\end{aligned}
\label{eqn:final_loss}
\end{equation}

\subsection{Training Details}
\label{subsec:details}

At stage one, we train the VQ-VAE model (Figure~\ref{fig:motion_prior}) on a single NVIDIA V100 for 200 epochs ($\sim$2 hours) with the AdamW~\cite{loshchilov2017decoupled} optimizer ($\beta_1=0.9$, $\beta_2=0.999$ and $\epsilon=1e-8$), where the learning rate is initialized as $10^{-4}$, and the mini-batch size is set as 1. At stage two, we train the temporal autoregressive model with the Adam optimizer~\cite{kingma2014adam}. The training duration is 100 epochs ($\sim$3 hours) and other hyper-parameters remain unchanged as stage one. 
\vspace{-8mm}
\paragraph{Style embedding space.}
The style embedding space is linearly spanned by $M$ learned basis vectors, where each style is represented by a style vector $\mathbf{s}$ that serves as the linear combination coefficients or a coordinate. In VOCA~\cite{cudeiro2019capture}, it assigns each speaker with a category-like one-hot style vector. Instead, we propose a concept to formulate a uniform style space via some learnable basis vectors (Eq.~\ref{eqn:embedding}), where the style vector is no longer bound with speaker ID but associated with every talking sample. During training, considering the limited style diversity of training datasets, we assign each speaker (\eg no. $i$) with a standard unit vector $\mathbf{e}_i$ as a style vector, under the assumption that each speaker is associated with a unique and consistent style. Anyway, arbitrary style vectors are allowed to interpolate new talking styles during inference.

\vspace{-2mm}
\section{Experiments}
\label{sec:exp}

\subsection{Datasets and Implementations}
We employ two widely used datasets, BIWI~\cite{fanelli20103} and VOCASET~\cite{cudeiro2019capture}, to train and test different methods in our experiments. Both datasets contain 4D face scans together with utterances spoken in English. BIWI contains 40 unique sentences shared across all speakers in the dataset, while VOCASET contains 255 unique sentences, which are partially shared among different speakers.

\textbf{BIWI dataset.}
BIWI is a 3D audio-visual corpus of affective speech and facial expression in the form of dense dynamic 3D face geometries, which is originally proposed to study affective communication. There is a total of 40 sentences uttered by 14 subjects, eight females and six males. Each sentence was recorded twice: with and without emotion. On average, each sentence is 4.67 seconds long. The 3D face dynamics are captured at 25fps, each with 23370 vertices and registered topology. We follow the data splits in~\cite{fan2022faceformer} and use the emotional subset. Specifically, the training set (BIWI-Train) contains 192 sentences, while the validation set (BIWI-Val) contains 24 sentences. There are two testing sets, in which BIWI-Test-A includes 24 sentences spoken by six seen subjects and BIWI-Test-B contains 32 sentences spoken by eight unseen subjects. BIWI-Test-A can be used for both quantitative and qualitative evaluation due to the seen subjects during training, while BIWI-Test-B is more suitable for qualitative evaluation.

\textbf{VOCASET dataset.}
VOCASET is comprised of 480 paired audio-visual sequences recorded from 12 subjects. The facial motion is captured at 60fps and is about 4 seconds long. Different from BIWI, each 3D face mesh is registered to the FLAME~\cite{li2017learning} topology with 5023 vertices. We adopt the same training (VOCA-Train), validation (VOCA-Val), and testing (VOCA-Test) splits as VOCA~\cite{cudeiro2019capture} and FaceFormer for fair comparisons.

\textbf{Implementations.}
We compare our work with three state-of-the-art methods: VOCA~\cite{cudeiro2019capture}, MeshTalk~\cite{richard2021meshtalk} and FaceFormer~\cite{fan2022faceformer}. We train and test VOCA on BIWI using the official codebase, while directly testing the released model that was trained on VOCASET. For MeshTalk, we train and test it using the official implementation on the two datasets. To compare with FaceFormer, we conduct testing directly using the pre-trained weights. Among the four methods, VOCA, FaceFormer and our CodeTalker require conditioning on a training speaking style during testing. For unseen subjects, we generate facial animations conditioned on all training styles. More details about the implementation details can be found in the \textit{Supplementary Material}.

\begin{table}[!t]
\centering
\setlength\tabcolsep{7pt}
\caption{
Quantitative evaluation on BIWI-Test-A. Lower means better for both metrics. 
}\vspace{-0.5em}
{
\resizebox{0.47\textwidth}{!}{
  \begin{tabular}{lcc}
  \toprule[0.8pt]
  \multirow{2}{*}{Method}   & Lip Vertex Error      & FDD \\
                            & ($\times10^{-4}$ mm)  & ($\times10^{-5}$ mm) \\ \hline\\[-2.2ex]
  VOCA                      & 6.5563                & 8.1816\\
  MeshTalk                  & 5.9181                & 5.1025\\ 
  FaceFormer                & 5.3077                & 4.6408\\
  CodeTalker (Ours)         & \textbf{4.7914}       & \textbf{4.1170}\\
  \bottomrule[0.8pt]
  \end{tabular}
  }
  }
  \label{tab:quantitative_eval}
  \vspace{-4mm}
\end{table}


\subsection{Quantitative Evaluation}
Following MeshTalk~\cite{richard2021meshtalk} and FaceFormer~\cite{fan2022faceformer}, we adopt the lip vertex error to measure the lip synchronization, which is the only publicly proposed metric for speech-driven facial animation evaluation, to our best knowledge. As a complement, we introduce a new quantitative measurement, \ie, upper-face motion statistics, to evaluate the overall facial dynamics.

\textbf{Lip vertex error.} It measures the lip deviation of a sequence with respect to the ground truth, \ie, calculating the maximal L2 error of all lip vertices for each frame and takes the average over all frames. 

\textbf{Upper-face dynamics deviation.}
The upper-face expression is just loosely correlated with the speech, depending on personal talking styles and the semantics of speech content. With this belief, we propose to measure the variation of facial dynamics for a motion sequence in comparison with that of the ground truth. Specifically, the \textit{upper-face dynamics deviation} (FDD) is calculated by:
\begin{equation}
\text{FDD}(\mathbf{M}_{1:T}, \hat{\mathbf{M}}_{1:T})=\frac{\sum_{v \in \mathcal{S}_U} (\text{dyn}(\mathbf{M}_{1:T}^v)-\text{dyn}(\hat{\mathbf{M}}_{1:T}^v))}{|\mathcal{S}_U|},
\end{equation}
where $\mathbf{M}_{1:T}^v \in \mathbb{R}^{3 \times T}$ denotes the motions of the $v$-th vertex, and $\mathcal{S}_U$ is the index set of upper-face vertices.
$\text{dyn}(\cdot)$ denotes the standard deviation of the element-wise L2 norm along the temporal axis.

We calculate the lip vertex error and upper-face dynamics deviation (FDD) over all sequences in BIWI-Test-A and take the average for comparison. According to Table~\ref{tab:quantitative_eval}, the proposed CodeTalker achieves lower error than the existing state-of-the-arts, suggesting that it produces more accurate lip-synchronized movements.
Besides, Table~\ref{tab:quantitative_eval} shows that our CodeTalker achieves the best performance in terms of FDD. 
%
It indicates the high consistency between the predicted upper-face expressions together with the trend of facial dynamics (conditioned on the speech and talking styles) and those of the ground truth.

\begin{figure*}[!t]
    \centering
    \includegraphics[width=1\linewidth]{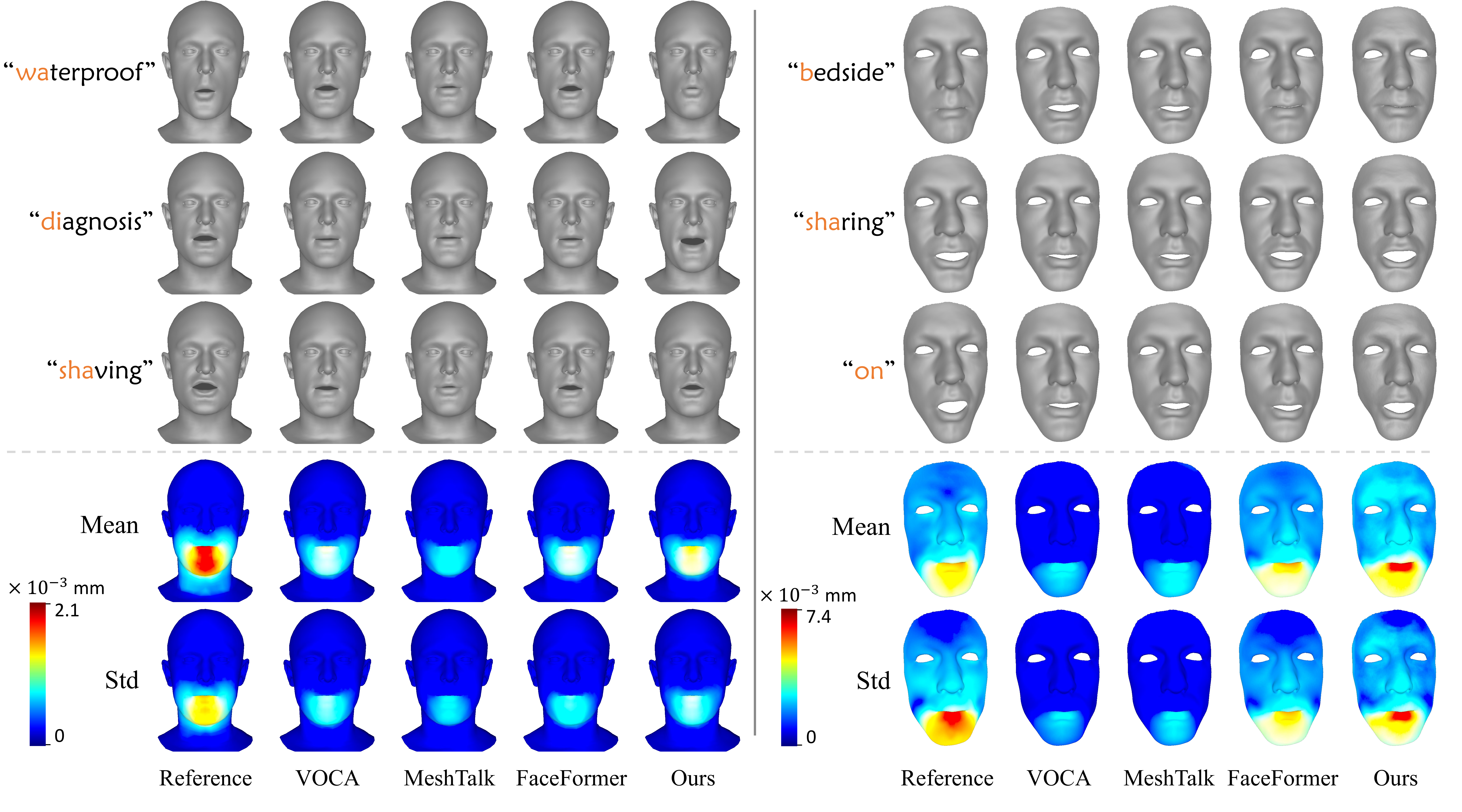} \vspace{-4mm}
    \caption{Visual comparisons of sampled facial motions animated by different methods on VOCA-Test (left) and BIWI-Test-B (right). The upper partition shows the facial animation conditioned on different speech parts, while the lower depicts the temporal statistics (mean and standard deviation) of adjacent-frame motion variations within a sequence.}
    \label{fig:sota_comparison}
    \vspace{-4mm}
\end{figure*}

\begin{figure}[!t]
    \centering
    \includegraphics[width=1\linewidth]{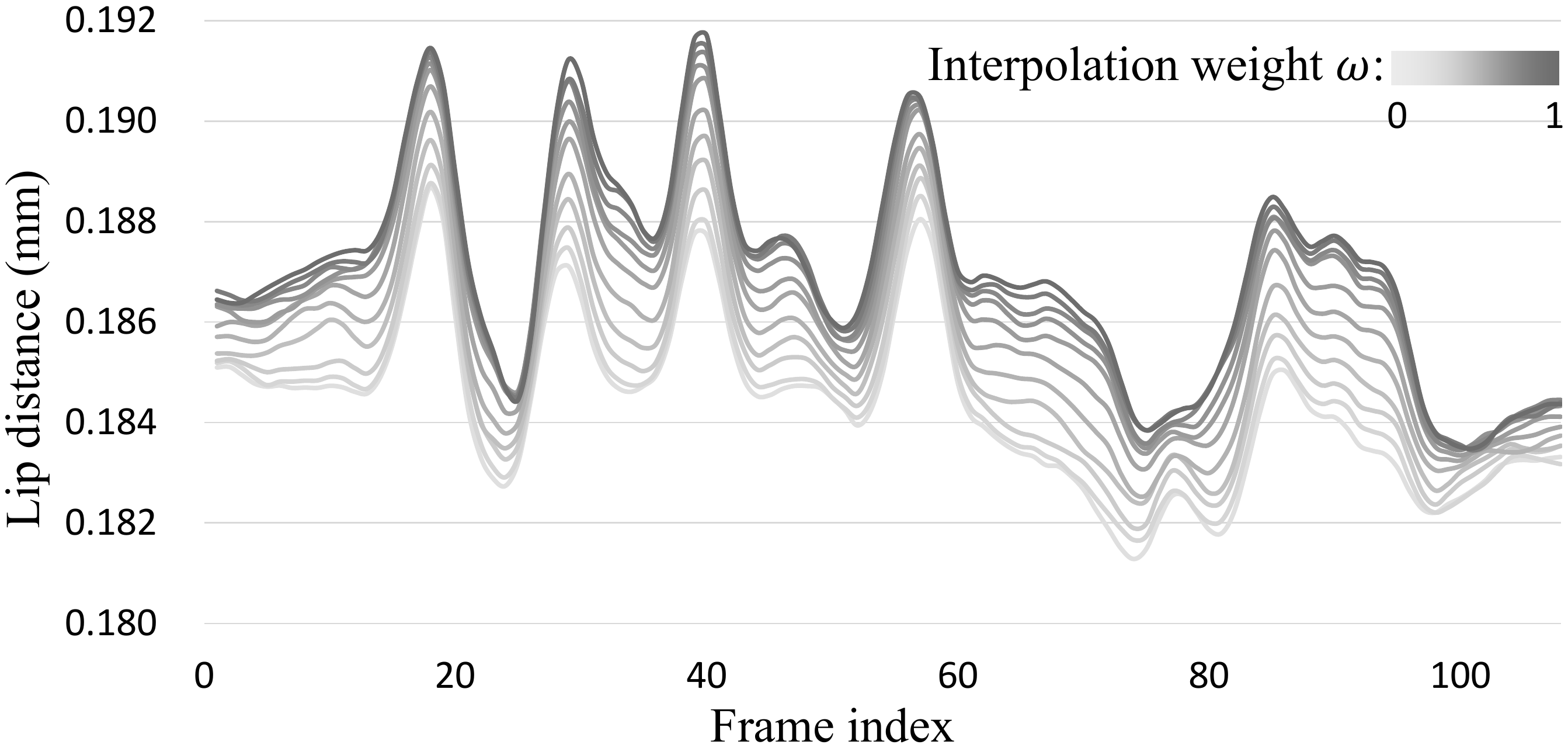}\vspace{-0.5em}
    \caption{Distance between lower and upper lip for our predictions within a sequence conditioned on different weighted linear combinations of two style vectors.}
    \label{fig:style_interpolation}
    \vspace{-5mm}
\end{figure}

\subsection{Qualitative Evaluation}
\label{subsec:qualitative}
We visually compare our method with other competitors in Figure~\ref{fig:sota_comparison}. For fair comparison, we assign the same talking style to VOCA, FaceFormer and our CodeTalker as conditional input, which is sampled at random.
To check the lip synchronization performance, we illustrate three typical frames of synthesized facial animations that speak at specific syllables, as compared in the upper partition in Figure~\ref{fig:sota_comparison}. We can observe that compared with the competitors, the lip movements produced by our CodeTalker are more accurately articulated with the speech signals and also more consistent with those of the reference. For example, CodeTalker produces better lip sync with proper mouth closures when pronouncing bilabial consonant /b/ (\ie, ``bedside" in the upper-right case of Figure~\ref{fig:sota_comparison}), compared to VOCA and MeshTalk; for the even challenging speech parts ``waterproof" and ``shaving" that need to pout, CodeTalker can produce accurate lip shapes while other methods suffer from the over-smoothing problem and fail to lip-sync correctly (Zoom in for better inspection).

Different from lip movements, facial expressions only have weak correlations with the speech signal, which tends to be static in front of cross-modal mapping ambiguity. To visualize the facial motion dynamics, we calculate the temporal statistics of adjacent-frame facial motions within a sequence. Specifically, we first calculate the inter-frame motion L2 distance and then compute the mean and standard deviation (std) across the sequence at each vertex.
The higher mean value indicates stronger facial movements, while the higher std value suggests richer variations of facial dynamics.
Two examples are visualized in the last two rows of Figure~\ref{fig:sota_comparison}, evidencing that our method outperforms others in achieving both stronger facial movements and a broader range of dynamics. It is mainly attributed to the superiority of the discrete facial motion space, which promotes the robustness to cross-modal uncertainty effectively.
Readers are recommended to watch the animation comparisons in the \textit{Supplemental Video}.

\textbf{Talking style interpolation.}
Our model can synthesize new speaking styles from the learned style embedding space.
To inspect the effects on VOCA-Test, we select two speaking style vectors, \ie, $\mathbf{e}_i$ and $\mathbf{e}_j$, which correspond to large and slight lip articulations respectively, and interpolate new talking style vectors $\mathbf{s}_\text{new}= \mathbf{B} \cdot [\omega \mathbf{e}_i + (1-\omega)\mathbf{e}_j]$ with a linear coefficient $\omega$. For the synthesized 3D animations of a sampled sequence, we plot the lower-upper lip distances across frames for each style  in Figure~\ref{fig:style_interpolation}, from which we observe the smooth transition of mouth amplitudes between the two typical styles.
It is not only useful to synthesize new talking styles, but also practical to match a specific speaking performance of an unseen subject during training.

\subsection{User Study}
The human perception system has been evolutionarily adapted to understanding subtle facial motions and capturing lip synchronization. Thus, it is still the most reliable measure in the speech-driven facial animation task. We conduct a user study to evaluate the quality of animated faces in perceptual lip synchronization and realism, compared with VOCA, MeshTalk, FaceFormer and the ground truth. We adopt A/B tests for each comparison, \ie, ours \emph{vs.} competitor, in terms of realistic facial animation and lip sync.
\begin{table}[!t]
\centering
\setlength\tabcolsep{3pt}
\caption{
  User study results on BIWI-Test-B and VOCA-Test. We adopt A/B testing and report the percentage of answers where A is preferred over B.
}\vspace{-0.5em}
\resizebox{0.48\textwidth}{!}{
\begin{tabular}{lcccc}
  \toprule[0.8pt] 
  \multirow{2}{*}{Competitors}   & \multicolumn{2}{c}{{ BIWI-Test-B}}    & \multicolumn{2}{c}{{ VOCA-Test}}    \\
  \cmidrule(lr){2-3} \cmidrule(lr){4-5}
                                & Lip Sync & Realism                 & Lip Sync & Realism \\
  \\[-2.5ex] \hline \\[-2.2ex]
  Ours \emph{vs.} VOCA          &  92.47& 89.25                       & 86.02&  84.95 \\
  Ours \emph{vs.} MeshTalk      & 80.65 & 82.80                       & 95.70 & 92.47  \\
  Ours \emph{vs.} FaceFormer    & 53.76 &  56.99                      &  70.97&  69.89 \\
  Ours \emph{vs.} GT            & 43.01 &  49.46                      &  43.01 & 43.01 \\
  \bottomrule[0.8pt]
  \end{tabular}
  }
  \label{tab:user_study}
  \vspace{-1.5em}
\end{table}
For BIWI, we obtain the results of four kinds of comparisons by randomly selecting 30 samples from BIWI-Test-B, respectively. To achieve the most variations in terms of speaking styles, we ensure the sampling results can fairly cover all conditioning styles. Thus, 120 A \emph{vs.} B pairs (30 samples $\times$ 4 comparisons) are created for BIWI-Test-B. Each pair is judged by at least 3 different participants separately, and finally, 372 entries are collected in total. For the user study on VOCASET, we apply the same setting as that on the BIWI dataset, \ie, another 120 A \emph{vs.} B pairs from VOCA-Test set, finally yielding 372 entries as well. In this study, 31 participants with good vision and hearing ability complete the evaluation successfully. Moreover, each participant is involved in all 8 kinds of comparisons to make better exposure and cover the diversity of favorability.

The percentage of A/B testing in terms of lip sync and realism on BIWI-Test-B is tabulated in Table~\ref{tab:user_study}, which shows that participants favor CodeTalker over competitors. Based on the visual analysis in Section~\ref{subsec:qualitative}, we attribute this to the facial animation synthesized by CodeTalker having more expressive facial motions, accurate lip shape, and well-synchronized mouth movements. For VOCA-Test, which has a nature of fewer upper-face motions, a similar favorability can still be observed in Table~\ref{tab:user_study}. We believe the reasons are at least three-fold: subtle motions around the eyes, more accurate lip movements and expressive motions in the lower face. Although be aware of a gap between our predictions and the recorded performance (ground truth), we surprisingly get over 40\% preference still. Overall, the user study justifies that the facial animations produced by CodeTalker have superior perceptual quality.

\begin{table}
\centering
\caption{
  Ablation study on the representation space of codebook. The performance is measured by the reconstruction error (\ie, L2 error) and lip vertex error on VOCA-Test and BIWI-Test-A.
}\vspace{-0.5em}
\resizebox{0.48\textwidth}{!}{
  \begin{tabular}{lccc}
  \toprule[0.8pt]
  \multirow{3}{*}{Variants} & VOCA-Test & \multicolumn{2}{c}{BIWI-Test-A} \\ \cmidrule(lr){2-2}\cmidrule(lr){3-4}
     & Rec. Error & Rec. Error & Lip Vertex Error \\
     & ($\times10^{-5}$ mm) & ($\times10^{-5}$ mm) & ($\times10^{-4}$ mm) \\
  \hline \\[-2.2ex]
  Shape-ent. codebook  & 2.75 & 4.07   & 6.41\\
  Motion codebook (Ours) &  \bf0.08 &  \bf2.83  &  \bf4.79\\
  \bottomrule[0.8pt]
  \end{tabular}
  }
  \label{tab:coding_geometry}
  \vspace{-0.5em}
\end{table}

\begin{figure}[!t]
\vspace{-0.4em}
    \centering
    \includegraphics[width=1\linewidth]{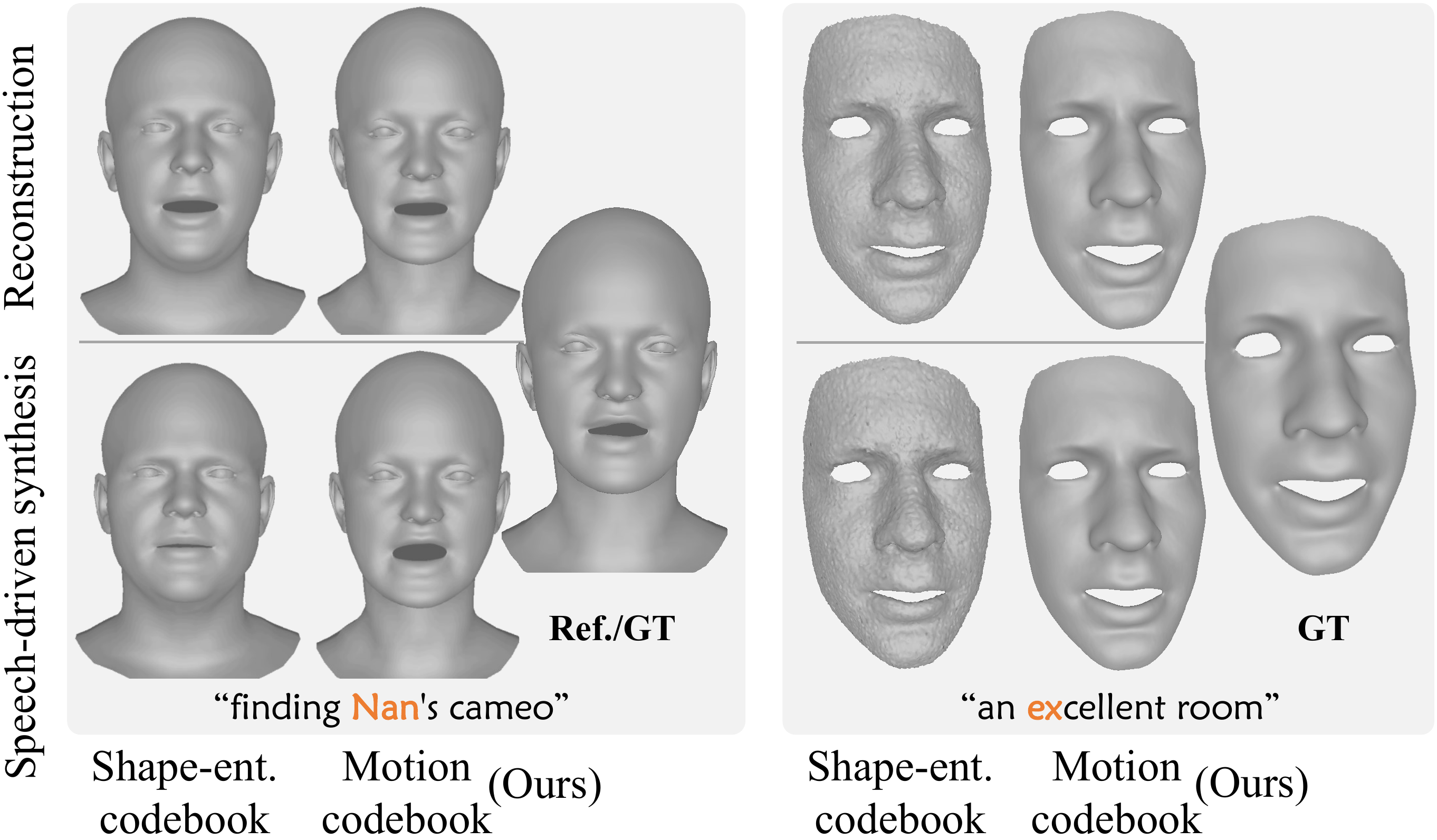}
    \caption{Visual comparisons of reconstruction and speech-driven motion synthesis results with different representation spaces on VOCA-Test (left) and BIWI-Test-A (right).}
    \label{fig:ablation_geo}
    \vspace{-1.5em}
\end{figure}

\subsection{Ablation Studies}
\label{subsec:ablation}
We study several key designs of our proposed method in this section, including the representation space and the hyper-parameters of the codebook construction.

\textbf{Representation space.}
To study the superiority of our motion-based representation, we construct a baseline that learns a shape-entangled codebook, \ie, the codes represent shapes ($\mathbf{m_t+h}$) instead of motions ($\mathbf{m_t}$).
As shown in Table~\ref{tab:coding_geometry}, the baseline decreases the reconstruction accuracy significantly, which is evidenced by the visualized examples in Figure~\ref{fig:ablation_geo}. It is mainly because the shape-entangled sequence contains more speaker-specific information that hinders the reusability of the codes. A direct weakness is the poor generalization for self-reconstruction, which further impedes cross-modal mapping correctness. In contrast, our proposed speaker-agnostic motion representation is more effective to represent generic motion priors shared across individuals, and hence promotes the quality of both self-reconstruction and speech-driven motion synthesis.

\begin{figure}[!t]
    \centering
    \includegraphics[width=1\linewidth]{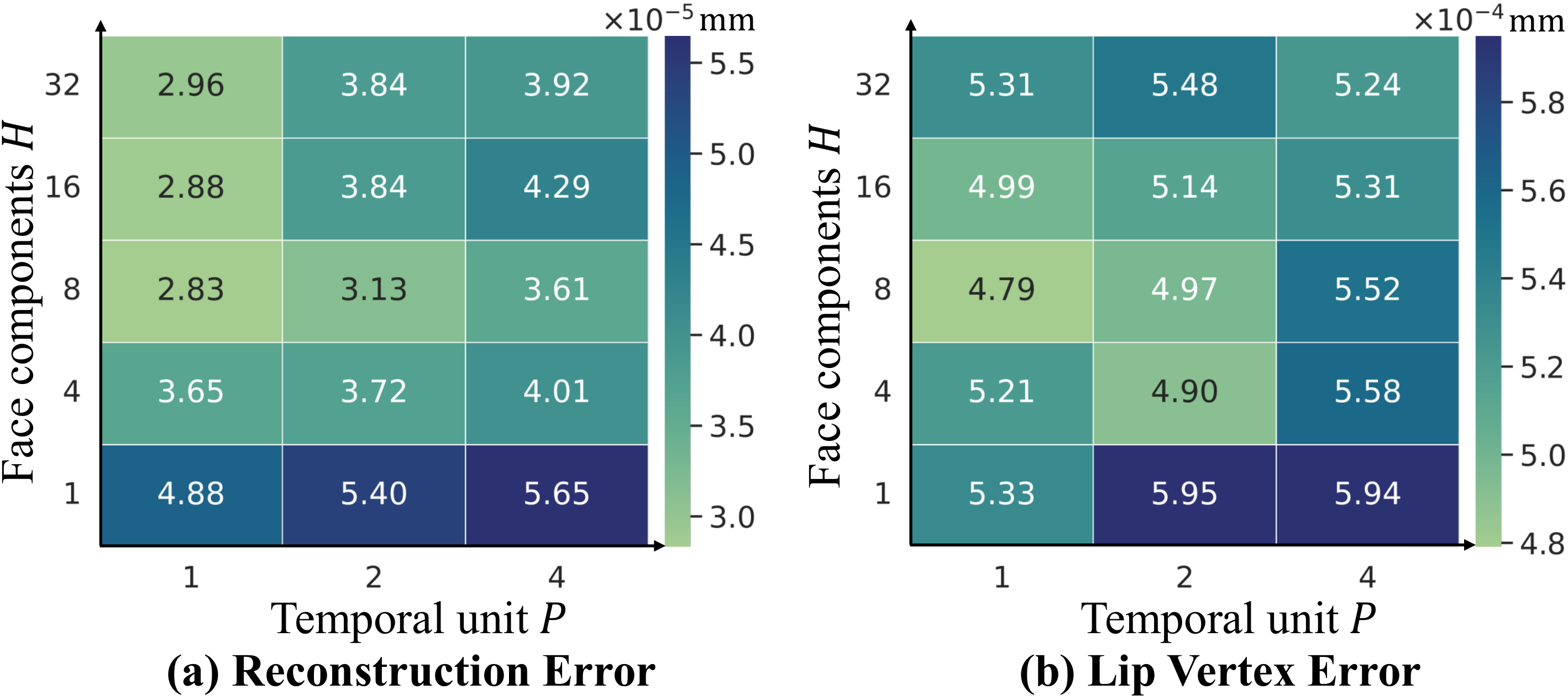}\vspace{-0.25em}
    \caption{Model performance comparisons with different hyper-parameters of the codebook, \ie, the length of temporal unit $P$ and the number of face components $H$. We measure the reconstruction error and lip vertex error on BIWI-Test-A.}
    \label{fig:ablation_codenumber}\vspace{-1.25em}
\end{figure}

\textbf{Codebook construction.}
We further study the hyper-parameters used for codebook construction. We evaluate the performance of different settings $\langle P,H\rangle$ by measuring their reconstruction accuracy and cross-modal mapping accuracy (namely lip vertex error).
First, we evaluate the reconstruction accuracy as shown in Figure~\ref{fig:ablation_codenumber}(a). On one hand, increasing $P$ degrades the reconstruction accuracy, which could be explained by the increased complexity of the motion to be represented. On the other hand, increasing $H$ eases the reconstruction but risks over-fitting, which explains the general benefits ($H < 8$) but inferior performance when $H\geq8$.
Notably, a similar trend could be found in the cross-modal mapping performance, as shown in Figure~\ref{fig:ablation_codenumber}(b). We conjecture that complex motion primitives cause lower reusability and higher redundancy, resulting in ambiguity in the cross-modal code query process.

\section{Discussion and Conclusion}
\label{sec:conclusion}

We demonstrated the advantages of casting speech-driven facial animation as a code query task in the discrete space, which notably promotes the motion synthesis quality against cross-modal ambiguity. By comparing to the existing state-of-the-arts, our proposed method shows superiority in achieving accurate lip sync and vivid facial expressions. However, we still follow the assumption that facial motions are independent of shapes, whose rationality may deserve further studies. Also, the overall perceptual quality still lags behind the ground truth, primarily due to the scarcity of paired audio-visual data. Additionally, the acquired generic motion prior adheres to the motion distribution delineated by the training set, which may deviate from real-world facial motions. As a future work, it is intriguing to guide the 3D facial animation by utilizing priors from large-scale available talking head videos.

\vspace{1mm}
\noindent\textbf{Acknowledgements. }This project is partially supported by Hong Kong Innovation and Technology Fund (ITF) (ref: ITS/307/20FP and ITS/313/20).

\newpage
{\small
\bibliographystyle{ieee_fullname}
\bibliography{egbib}
}

\clearpage

\appendix

\noindent{\Large \textbf{{Appendix}}}
\vspace{0.2mm}
\\

This supplemental document contains four sections: Section~\ref{sec:implementation} shows implementation details of our CodeTalker; Section~\ref{sec:more} presents more discussions on the proposed method; Section~\ref{sec:user} presents details of the user study; and Section~\ref{sec:video} presents short descriptions of the supplemental video. The source code and trained model will also be released upon publication.

\section{Implementation Details}
\label{sec:implementation}
\subsection{{Hyper-parameters of Codebook}}
We have explored and discussed the important hyper-parameters of our motion codebook in Section~\ref{subsec:ablation} ``Codebook construction" on the BIWI dataset in the main paper. Here we provide more specific parameters adopted for CodeTalker trained on the two datasets. For BIWI, we have the ground truth for quantitative evaluation on the testing set BIWI-Test-A to determine a group of parameters $P=1$ and $H=8$ for high-quality results (\ie, Section~\ref{subsec:ablation} ``Codebook construction" in the main paper). Additionally, we set the codebook item number $N=256$ and the dimension of items $C=128$. Although more codebook items and dimensions could ease reconstruction, the redundant elements may cause ambiguity in speech-driven motion synthesis. Hence, we did not heavily tune these parameters and just empirically set them for good visual quality. For VOCASET, since there is no ground truth for us to obtain the quantitative results, we empirically select a group of parameters (\ie, $N=256$, $P=1$, $H=16$, $C=64$), which could produce visually plausible facial animations in our experiments.

\subsection{Network Architecture}
To improve the reproducibility of our CodeTalker, we further illustrate the detailed network architectures for the facial motion space learning and the speech-driven motion synthesis (Section~\ref{subsec:motion_space} and~\ref{subsec:motion_prediction} in the main paper, respectively), which are shown Table~\ref{tab:network_arch}.

\section{More Discussions on CodeTalker}
\label{sec:more}
\begin{table*}[!t]
\centering
\caption{
Parameter illustration of network architectures. C(k,s,p,n) denotes a 1D Convolutional layer with kernel size k, stride size s, padding size p, and output channels of n. $\text{T}_\text{enc}$(d1,d2,h,l) denotes a transformer encoder layer with basic channel number of d1, forward channel number of d2, self-attention head number of h, and layer number l, while similarly, $\text{T}_\text{dec}$ represents a transformer decoder layer. L(n) denotes a linear layer with output channels of n. CA[$\cdot$] stands for the additional cross-attention input for transformer decoders. $\text{l}_\text{CM}=12$ for BIWI, while $\text{l}_\text{CM}=6$ for VOCASET. $n \cdot T$ stands for the interpolated audio feature length in order to align with visual frames, where $n=2$ for BIWI and $n=1$ for VOCASET. `+' denotes the channel-wise addition. ``Drop" means the dropout operation.
}
{
\resizebox{1\textwidth}{!}{
{\renewcommand{\arraystretch}{1.4}%
  \begin{tabular}{c|c|l|l}
  \toprule[0.8pt]
  Stage & Module  & Input $\rightarrow$ Output& Layer Operation \\
  \hline
  \multirow{7}{*}{I}&\multirow{4}{*}{Encoder}  & $\mathbf{M}(T, V,3) \rightarrow \mathbf{M}(T, V \cdot 3)$    &  Reshape\\
  &&    $\mathbf{M}(T,V \cdot 3) \rightarrow \mathbf{Z}^1_e(T, 1024)$   & L(1024) $\rightarrow$ LReLU $\rightarrow$ C(5,1,2,1024) $\rightarrow$ LReLU $\rightarrow$ IN \\ 
  &&    $\mathbf{Z}^1_e(T, 1024) \rightarrow \mathbf{Z}^2_e(T,H \cdot C)$    & L(1024) $\rightarrow$ $\text{T}_\text{enc}$(1024,1536,8,6) $\rightarrow$ L($H \cdot C$)\\
  && $\mathbf{Z}^2_e(T,H \cdot C) \rightarrow \mathbf{Z}_\Quantize(T,H,C)$ & Reshape $\rightarrow$ Quantize \\
  \cline{2-4}
  &\multirow{3}{*}{Decoder}    & $\mathbf{Z}_\Quantize(T,H,C) \rightarrow \mathbf{Z}_\Quantize(T,H \cdot C)$     & Reshape\\
  &&$\mathbf{Z}_\Quantize(T,H \cdot C) \rightarrow \mathbf{Z}^1_d(T,1024)$ & L(1024) $\rightarrow$ C(5,1,2,1024) $\rightarrow$ LReLU $\rightarrow$ IN\\
  && $\mathbf{Z}^1_d(T,1024) \rightarrow \mathbf{\hat{M}}(T,V \cdot 3)$& L(1024) $\rightarrow$ $\text{T}_\text{enc}$(1024,1536,8,6) $\rightarrow$ L($V \cdot 3$)\\
  \hline
  \multirow{12}{*}{II}& & \multirow{4}{*}{$\mathbf{A}(T, d) \rightarrow \mathbf{F}_e^1(T^\prime, 512)$} &    C(10,5,0,512) $\rightarrow$ GN $\rightarrow$ GeLU $\rightarrow$ C(3,2,0,512) $\rightarrow$ GN $\rightarrow$ GeLU \\
  &&&  $\rightarrow$ C(3,2,0,512) $\rightarrow$ GN $\rightarrow$ GeLU $\rightarrow$ C(3,2,0,512) $\rightarrow$ GN $\rightarrow$ GeLU\\
  &Speech&&  $\rightarrow$ C(3,2,0,512) $\rightarrow$ GN $\rightarrow$ GeLU $\rightarrow$ C(3,2,0,512) $\rightarrow$ GN $\rightarrow$ GeLU\\
  &Encoder&&  $\rightarrow$ C(2,2,0,512) $\rightarrow$ GN $\rightarrow$ GeLU $\rightarrow$ C(2,2,0,512) $\rightarrow$ GN $\rightarrow$ GeLU\\
  &&   $\mathbf{F}_e^1(T^\prime, 512) \rightarrow \mathbf{F}_e^2(n \cdot T, 768)$  & Interpolate $\rightarrow$ LN $\rightarrow$ L(768) $\rightarrow$ Drop \\
  && $\mathbf{F}_e^2(n \cdot T, 768) \rightarrow \mathbf{F}_e^3(n \cdot T, 1024)$  &  $\text{T}_\text{enc}$(768,3072,12,12) $\rightarrow$ L(1024) \\
  \cline{2-4}
  &&  $\mathbf{\hat{M}}_\text{past}(T,V \cdot 3) \rightarrow \mathbf{F}_\text{emb}^\text{past}(T, 1024)$ & L(1024) $\rightarrow$ +StyleVector \\
  && $\mathbf{F}_\text{emb}^\text{past}(T, 1024) \rightarrow \mathbf{\hat{Z}}_d^1(T, 1024)$ &  $\text{T}_\text{dec}$(1024,2048,4,$\text{l}_\text{CM}$) with CA[$\mathbf{F}_e^3$] $\rightarrow$ L($H \cdot C$) \\
  &Cross-modal& $\mathbf{\hat{Z}}_d^1(T, H \cdot C) \rightarrow \mathbf{\hat{Z}}_\Quantize(T,H,C)$ &  Reshape $\rightarrow$ Quantize  \\
  & Decoder  & $\mathbf{\hat{Z}}_\Quantize(T,H,C) \rightarrow \mathbf{\hat{Z}}_\Quantize(T,H \cdot C)$     & Reshape\\
  &&$\mathbf{\hat{Z}}_\Quantize(T,H \cdot C) \rightarrow \mathbf{\hat{Z}}^2_d(T,1024)$ & L(1024) $\rightarrow$ C(5,1,2,1024) $\rightarrow$ LReLU $\rightarrow$ IN\\
  && $\mathbf{\hat{Z}}^2_d(T,1024) \rightarrow \mathbf{\hat{M}}(T,V \cdot 3)$& L(1024) $\rightarrow$ $\text{T}_\text{enc}$(1024,1536,8,6) $\rightarrow$ L($V \cdot 3$)\\
  \bottomrule[0.8pt]
  \end{tabular} 
    }
    }
  }
  \label{tab:network_arch}
\end{table*}

\begin{figure}[!t]
    \centering
    \includegraphics[width=1\linewidth]{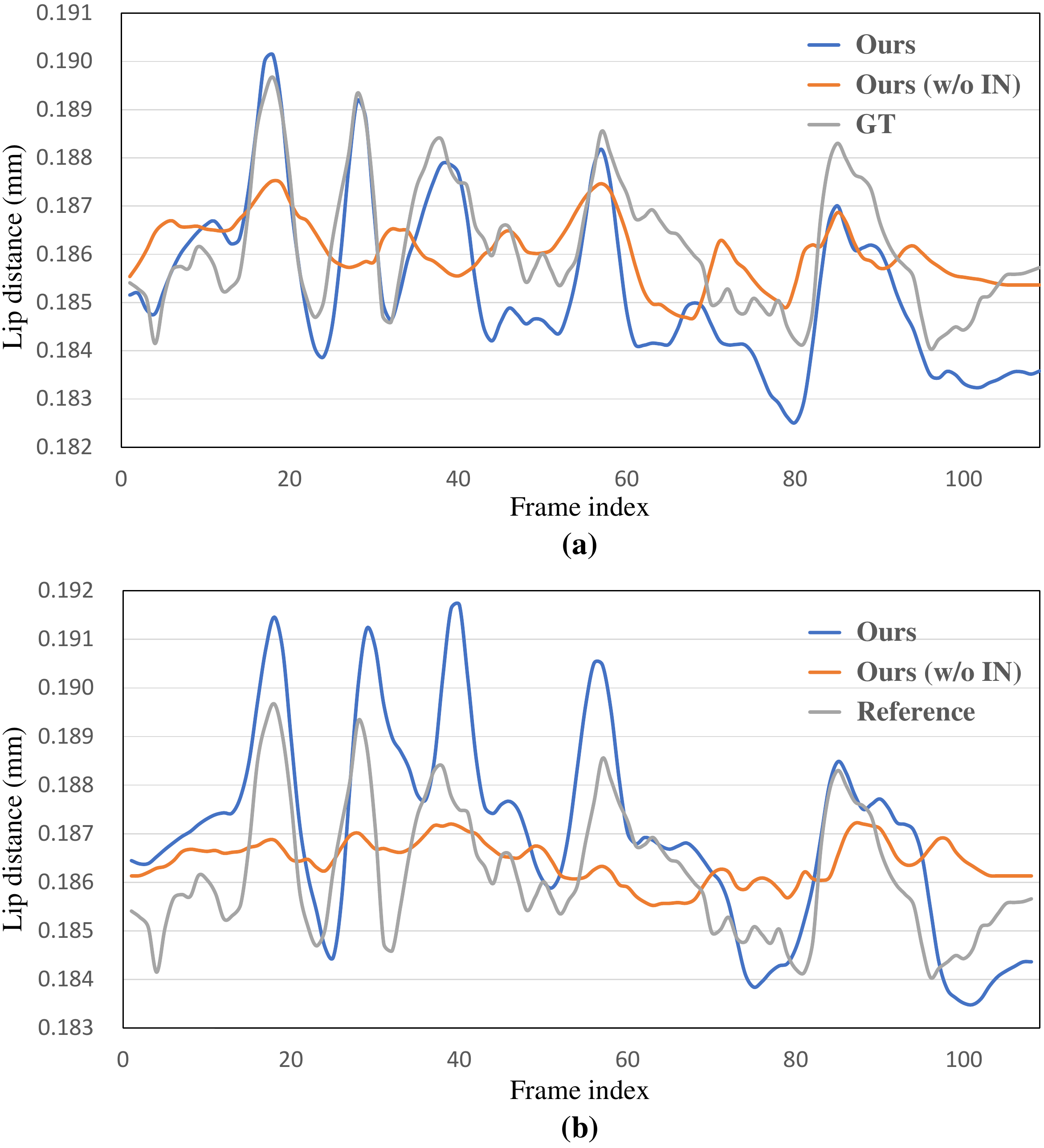}
    \caption{Distance between lower and upper lip within a sampled sequence from VOCA-Test of (a) reconstruction and (b) speech-driven motion synthesis results produced by different variants.}
    \label{fig:ablation_IN}
\end{figure}

\begin{table}
\centering
\caption{
  Ablation study on the Instance Normalization (IN) for self-reconstruction learning. The performance is measured by the reconstruction error on VOCA-Test and BIWI-Test-A.
}\vspace{-0.5em}
\resizebox{0.48\textwidth}{!}{
  \begin{tabular}{lccc}
  \toprule[0.8pt]
  \multirow{2}{*}{Variants} & \multicolumn{2}{c}{Reconstruction Error} \\ 
  \cmidrule{2-3}
     &  VOCA-Test ($\times10^{-5}$ mm) & BIWI-Test-A ($\times10^{-5}$ mm)\\
  \hline \\[-2.2ex]
  Ours (w/o IN)  & 0.12     & 3.27  \\
  Ours           &  \bf0.08 &  \bf2.83  \\
  \bottomrule[0.8pt]
  \end{tabular}
  }
  \label{tab:ablation_IN}
\end{table}

\subsection{Instance Normalization in Self-reconstruction Learning}
Instance Normalization~\cite{ulyanov2016instance} (IN) has been widely used in the filed of style transfer~\cite{ulyanov2017improved,huang2017arbitrary}, which is defined as:
\begin{equation}
    \text{IN}(x) = \gamma (\frac{x-\mu(x)}{\sigma(x)})+\beta.
\end{equation}
Different from BN~\cite{ioffe2015batch} layers, here $\mu(x)$ and $\sigma(x)$ are computed across temporal dimensions independently for each channel within each sample:
\begin{equation}
    \mu_{nc}(x) = \frac{1}{T}\sum_{t=1}^{T}x_{nct}
\end{equation}
\begin{equation}
    \sigma_{nc}(x) = \sqrt{\frac{1}{T}\sum_{t=1}^{T}(x_{nct}-\mu_{nc}(x))^2+\epsilon}
\end{equation}
Interestingly, we empirically find that normalizing feature statistics (\ie, mean and variance) with IN (not BN due to small mini-batch size) can boost the performance of our CodeTalker in self-reconstruction learning, as shown in Table~\ref{tab:ablation_IN}. In addition, it can also make self-reconstruction training more stable. To better show the gain of normalization, we also visualize the lip distance of a sampled sequence of reconstruction results from VOCA-Test in Figure~\ref{fig:ablation_IN}(a). The visualization result indicates that the predicted lip amplitudes are closer to those of the ground truth by equipping with IN, while the ablated variant (\ie, Ours (w/o IN)) cannot reconstruct lip movements with accurate amplitudes. The speech-driven facial motion synthesis (stage two) can also benefit from the facial motion codebook learned in self-reconstruction with IN, as shown in Figure~\ref{fig:ablation_IN}(b). Note that we synthesize facial motions conditioned on a randomly sampled speaking style. We conjecture that facial motions with different magnitudes could be well encapsulated into the discrete motion prior by normalizing temporal elements within each channel. The rationality and effect of IN deserve further studies as our potential direction.




\begin{table}[!t]
\centering
\setlength\tabcolsep{7pt}
\caption{
Comparison of lip-sync errors. We compare different methods on BIWI-Test-A. Lower means better. $\lambda$ is the weighting factor.
}\vspace{-0.5em}
{
\resizebox{0.48\textwidth}{!}{
  \begin{tabular}{lcc}
  \toprule[0.8pt]
  Method   & Lip Vertex Error ($\times10^{-4}$ mm)     \\ \hline\\[-2.2ex]
  Alter. ($\mathcal{L}_\text{ce}$)                            & 9.6356                      \\
  Alter. ($\lambda \mathcal{L}_\text{ce}$+$\mathcal{L}_\text{reg}$) & 5.1138                      \\
  Alter. ($\lambda \mathcal{L}_\text{ce}$+$\mathcal{L}_\text{reg}$+$\mathcal{L}_\text{motion}$) & 5.0254  \\ 
  CodeTalker (Ours)                                             & \textbf{4.7914}       \\
  \bottomrule[0.8pt]
  \end{tabular}
  }
  }
  \label{tab:dataflow}
  \vspace{-2mm}
\end{table}

\subsection{Alternative Data Flow and Supervision}
\begin{figure}[!t]
    \centering
    \includegraphics[width=1\linewidth]{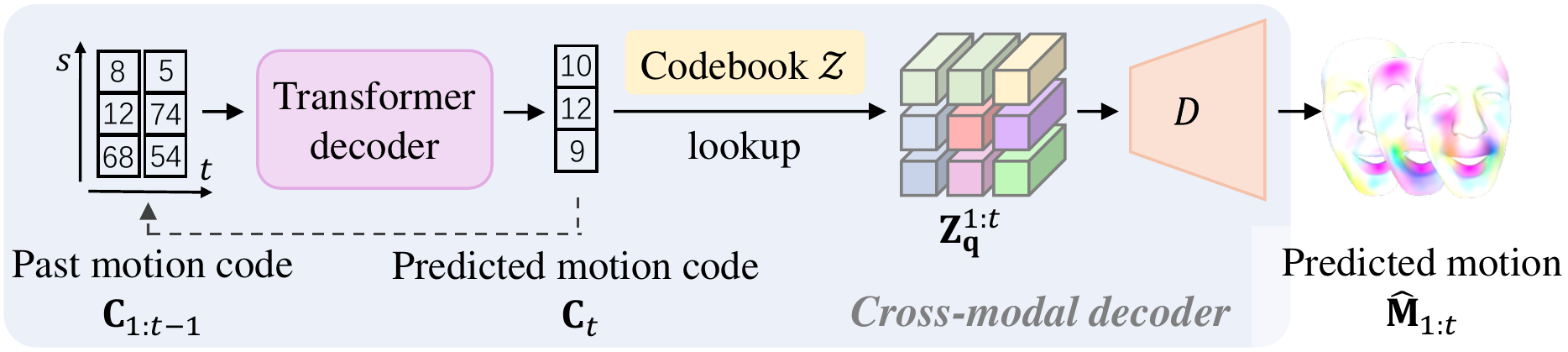}
    \caption{Alternative data flow and supervision framework of our cross-modal decoder. Note that we omit the style vector and audio features input for simplicity. Given the past motion code as input, the alternative cross-modal decoder first autoregressively predict motion code and then decode them into motions with the pre-trained codebook and decoder.}
    \label{fig:ablation_dataflow}
\end{figure}

As we have summarized in Section~\ref{subsec:discrete_prior_learning} of the main paper, recent works explore the power of discrete prior learning in a large variety of tasks, among which most existing Vector Quantization (VQ)-based works~\cite{ng2022learning,zhou2022towards} adopt categorical cross-entropy (CE) loss to supervise their token predictions. Hence, we also explore some alternative data flow and supervision frameworks as our cross-modal decoder, which is shown in Figure~\ref{fig:ablation_dataflow}. It is worth noting that the style vector and audio features are omitted for simplicity.

Different from our cross-modal decoder in the main paper, the alternative takes past motion code as input and then autoregressively predicts code sequences in form of n-way classification. The predicted code sequence then retrieves the respective code items from the learned codebook $\mathcal{Z}$, and further produces facial motion sequences through the fixed decoder $D$. A CE loss is adopted to penalize error between the predicted code sequence $\hat{\mathbf{c}}\in \{0,\dots,|N|-1\}^{T^\prime \cdot H}$ and the ground truth $\mathbf{c}$ generated by the pre-trained encoder $E$:
\begin{equation}
    \mathcal{L}_\text{ce}=\sum_{i=0}^{T^\prime \cdot H}-\mathbf{c}_i\log(\hat{\mathbf{c}}_i).
\end{equation}
We train the alternatives with the same settings as those in the main paper (Section~\ref{subsec:details}). The lip-sync evaluation result is tabulated in Table~\ref{tab:dataflow}. Alternative model with $\mathcal{L}_\text{ce}$ alone cannot converge well due to the difficult cross-modality mapping of token prediction. While adding more constraints (\ie, $\mathcal{L}_\text{reg}$ and $\mathcal{L}_\text{motion}$ in the main paper Eq.~\ref{eqn:final_loss}) can ease the difficulty of token prediction learning, the performance is still limited with this token prediction framework. Overall, the lower average lip error achieved by our CodeTalker suggests its framework superiority in terms of the accuracy of lip movements.

\footnotetext[1]{~\url{https://doubiiu.github.io/projects/codetalker}}

\section{User Study}
\label{sec:user}
\begin{figure*}[!t]
    \centering
    \includegraphics[width=0.85\linewidth, trim=0 0 0 1.7mm, clip]{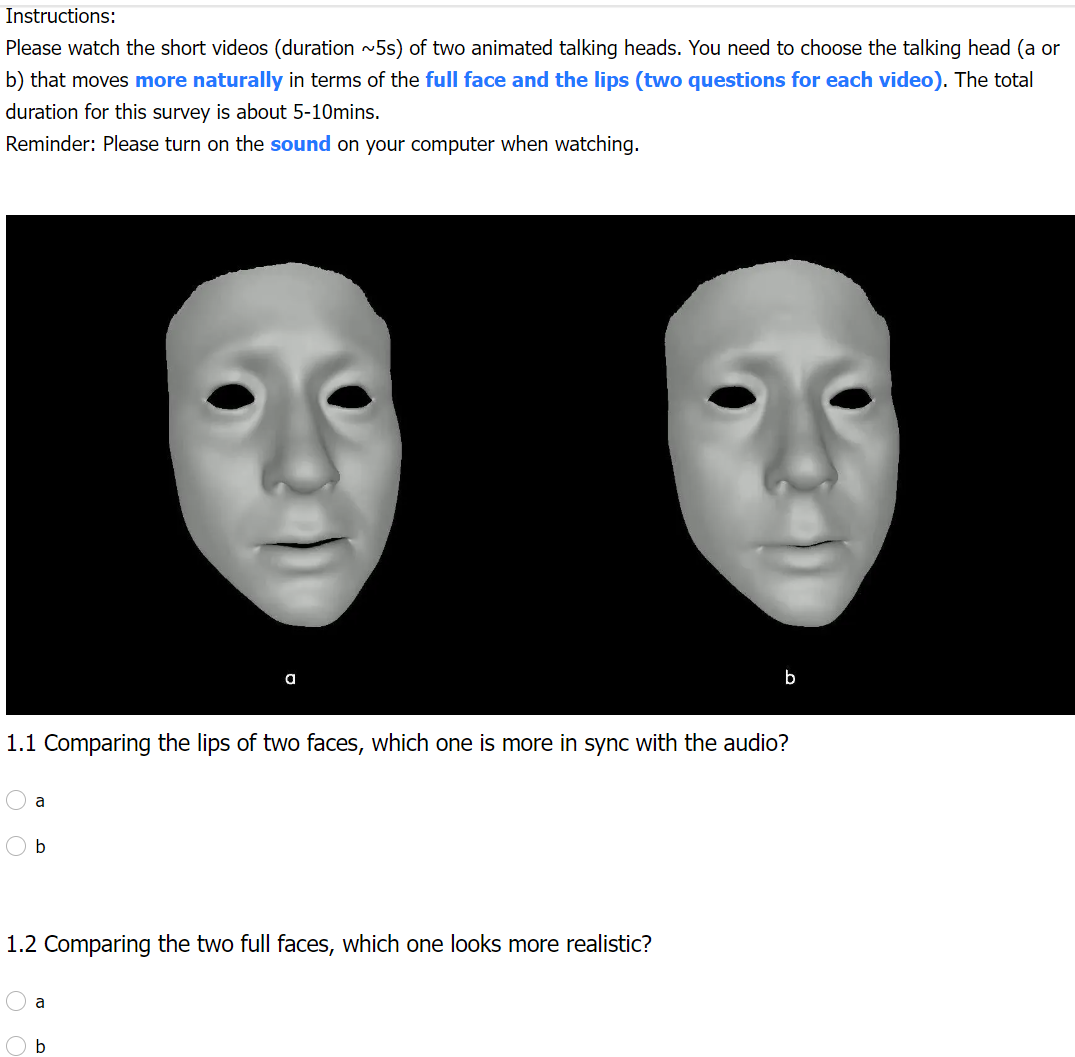}
    \caption{Designed user study interface. Each participant need to answer 24 video pairs and here only one video pair is shown due to the page limit.}
    \label{fig:user_study}
\end{figure*}

The designed user study interface is shown in Figure~\ref{fig:user_study}. A user study is expected to be completed with 5--10 minutes (24 video pairs $\times$ 5 seconds $\times$ 3 times watching). To remove the impact of random selection, we filter out those comparison results completed in less than two minutes. For each participant, the user study interface shows 24 video pairs and the participant is instructed to judge the videos twice with the following two questions, respectively: ``Comparing the lips of two faces, which one is more in sync with the audio?" and ``Comparing the two full faces, which one looks more realistic?".

\section{Video Comparison}
\label{sec:video}
To better evaluate the qualitative results produced by competitors~\cite{richard2021meshtalk,cudeiro2019capture,fan2022faceformer,karras2017audio} and our CodeTalker, we provide a supplemental video\footnotemark[1] for demonstration and comparison. Specifically, we test our model using various audio clips, including the audio clips extracted from TED and TEDx videos, audio sequences from the VOCASET and BIWI datasets, and the speech from supplementary videos of previous methods. The video shows that CodeTalker can synthesize natural and plausible facial animations with well-synchronized lip movements. It is worth noting that, compared to the competitors (\ie, VOCA, MeshTalk and FaceFormer) suffering from the over-smoothing problem, our CodeTalker can produce more vivid and realistic facial motions and better lip sync. Besides, we also show the talking style interpolation results and facial animations of talking in different languages.

\end{document}